\documentclass[conference]{IEEEtran}
\IEEEoverridecommandlockouts
\usepackage{amsmath,amssymb,amsfonts}
\usepackage{graphicx}
\usepackage{textcomp}
\usepackage{xcolor}
\usepackage{booktabs}
\usepackage{algorithm}
\usepackage{hyperref}
\usepackage{cleveref}
\usepackage{pgfplots}
\usepgfplotslibrary{groupplots}
\usepackage[noend]{algpseudocode}

\usepackage{siunitx}
\pgfplotsset{compat=newest}
\pgfplotsset{every axis legend/.append style={legend cell align=left}}

\pgfplotsset{every axis/.append style={
                    title style={font=\small},
                    tick label style={font=\footnotesize}  
                    }}
\pgfplotsset{every axis label/.style={font=\small}}                    
\pgfplotsset{
legend image code/.code={
\draw[mark repeat=2,mark phase=2]
plot coordinates {
(0cm,0cm)
(0.15cm,0cm)        %% default is (0.3cm,0cm)
(0.3cm,0cm)         %% default is (0.6cm,0cm)
};%
}
}

\definecolor{pastelMagenta}{HTML}{FF48CF}
\definecolor{pastelPurple}{HTML}{8770FE}
\definecolor{pastelBlue}{HTML}{1BA1EA}
\definecolor{pastelSeaGreen}{HTML}{14B57F}
\definecolor{pastelGreen}{HTML}{3EAA0D}
\definecolor{pastelOrange}{HTML}{C38D09}
\definecolor{pastelRed}{HTML}{F5615C}

\definecolor{ra_1h}{rgb}{1.0, 1.0, 1.0}
\definecolor{ra_2h}{rgb}{1.0, 0.5019607843137255, 0.0}
\definecolor{ra_3h}{rgb}{0.0, 0.6, 0.0}

\definecolor{ra_1}{rgb}{1.0, 1.0, 1.0}
\definecolor{ra_2}{rgb}{0.00392156862745098, 0.34509803921568627, 0.792156862745098}
\definecolor{ra_3}{rgb}{0.7450980392156863, 0.0, 0.0}

\pgfplotsset{
   	colormap={pasteljet}{
		rgb=(0.99325,0.90616,0.14394)
		rgb=(0.98387,0.90487,0.13690)
		rgb=(0.97442,0.90359,0.13021)
		rgb=(0.96489,0.90232,0.12394)
		rgb=(0.95530,0.90107,0.11813)
		rgb=(0.94564,0.89982,0.11284)
		rgb=(0.93590,0.89857,0.10813)
		rgb=(0.92611,0.89733,0.10407)
		rgb=(0.91624,0.89609,0.10072)
		rgb=(0.90631,0.89485,0.09813)
		rgb=(0.89632,0.89362,0.09634)
		rgb=(0.88627,0.89237,0.09537)
		rgb=(0.87617,0.89112,0.09525)
		rgb=(0.86601,0.88987,0.09595)
		rgb=(0.85581,0.88860,0.09745)
		rgb=(0.84556,0.88732,0.09970)
		rgb=(0.83527,0.88603,0.10265)
		rgb=(0.82494,0.88472,0.10622)
		rgb=(0.81458,0.88339,0.11035)
		rgb=(0.80418,0.88205,0.11496)
		rgb=(0.79376,0.88068,0.12001)
		rgb=(0.78331,0.87928,0.12540)
		rgb=(0.77285,0.87787,0.13111)
		rgb=(0.76237,0.87642,0.13706)
		rgb=(0.75188,0.87495,0.14323)
		rgb=(0.74139,0.87345,0.14956)
		rgb=(0.73089,0.87192,0.15603)
		rgb=(0.72039,0.87035,0.16260)
		rgb=(0.70990,0.86875,0.16926)
		rgb=(0.69942,0.86712,0.17597)
		rgb=(0.68894,0.86545,0.18272)
		rgb=(0.67849,0.86374,0.18950)
		rgb=(0.66805,0.86200,0.19629)
		rgb=(0.65764,0.86022,0.20308)
		rgb=(0.64726,0.85840,0.20986)
		rgb=(0.63690,0.85654,0.21662)
		rgb=(0.62658,0.85464,0.22335)
		rgb=(0.61629,0.85271,0.23005)
		rgb=(0.60604,0.85073,0.23671)
		rgb=(0.59584,0.84872,0.24333)
		rgb=(0.58568,0.84666,0.24990)
		rgb=(0.57556,0.84457,0.25642)
		rgb=(0.56550,0.84243,0.26288)
		rgb=(0.55548,0.84025,0.26928)
		rgb=(0.54552,0.83804,0.27563)
		rgb=(0.53562,0.83579,0.28191)
		rgb=(0.52578,0.83349,0.28813)
		rgb=(0.51599,0.83116,0.29428)
		rgb=(0.50627,0.82879,0.30036)
		rgb=(0.49661,0.82638,0.30638)
		rgb=(0.48703,0.82393,0.31232)
		rgb=(0.47750,0.82144,0.31820)
		rgb=(0.46805,0.81892,0.32400)
		rgb=(0.45867,0.81636,0.32973)
		rgb=(0.44937,0.81377,0.33538)
		rgb=(0.44014,0.81114,0.34097)
		rgb=(0.43098,0.80847,0.34648)
		rgb=(0.42191,0.80577,0.35191)
		rgb=(0.41291,0.80304,0.35727)
		rgb=(0.40400,0.80027,0.36255)
		rgb=(0.39517,0.79748,0.36776)
		rgb=(0.38643,0.79464,0.37289)
		rgb=(0.37778,0.79178,0.37794)
		rgb=(0.36921,0.78889,0.38291)
		rgb=(0.36074,0.78596,0.38781)
		rgb=(0.35236,0.78301,0.39264)
		rgb=(0.34407,0.78003,0.39738)
		rgb=(0.33588,0.77702,0.40205)
		rgb=(0.32780,0.77398,0.40664)
		rgb=(0.31981,0.77091,0.41115)
		rgb=(0.31193,0.76782,0.41559)
		rgb=(0.30415,0.76470,0.41994)
		rgb=(0.29648,0.76156,0.42422)
		rgb=(0.28892,0.75839,0.42843)
		rgb=(0.28148,0.75520,0.43255)
		rgb=(0.27415,0.75199,0.43660)
		rgb=(0.26694,0.74875,0.44057)
		rgb=(0.25986,0.74549,0.44447)
		rgb=(0.25290,0.74221,0.44828)
		rgb=(0.24607,0.73891,0.45202)
		rgb=(0.23937,0.73559,0.45569)
		rgb=(0.23281,0.73225,0.45928)
		rgb=(0.22640,0.72889,0.46279)
		rgb=(0.22012,0.72551,0.46623)
		rgb=(0.21400,0.72211,0.46959)
		rgb=(0.20803,0.71870,0.47287)
		rgb=(0.20222,0.71527,0.47608)
		rgb=(0.19657,0.71183,0.47922)
		rgb=(0.19109,0.70837,0.48228)
		rgb=(0.18578,0.70489,0.48527)
		rgb=(0.18065,0.70140,0.48819)
		rgb=(0.17571,0.69790,0.49103)
		rgb=(0.17095,0.69438,0.49380)
		rgb=(0.16638,0.69086,0.49650)
		rgb=(0.16202,0.68732,0.49913)
		rgb=(0.15785,0.68376,0.50169)
		rgb=(0.15389,0.68020,0.50417)
		rgb=(0.15015,0.67663,0.50659)
		rgb=(0.14662,0.67305,0.50894)
		rgb=(0.14330,0.66946,0.51121)
		rgb=(0.14021,0.66586,0.51343)
		rgb=(0.13734,0.66225,0.51557)
		rgb=(0.13469,0.65864,0.51765)
		rgb=(0.13227,0.65501,0.51966)
		rgb=(0.13007,0.65138,0.52161)
		rgb=(0.12809,0.64775,0.52349)
		rgb=(0.12633,0.64411,0.52531)
		rgb=(0.12478,0.64046,0.52707)
		rgb=(0.12344,0.63681,0.52876)
		rgb=(0.12231,0.63315,0.53040)
		rgb=(0.12138,0.62949,0.53197)
		rgb=(0.12064,0.62583,0.53349)
		rgb=(0.12008,0.62216,0.53495)
		rgb=(0.11970,0.61849,0.53635)
		rgb=(0.11948,0.61482,0.53769)
		rgb=(0.11942,0.61114,0.53898)
		rgb=(0.11951,0.60746,0.54022)
		rgb=(0.11974,0.60379,0.54140)
		rgb=(0.12009,0.60010,0.54253)
		rgb=(0.12057,0.59642,0.54361)
		rgb=(0.12115,0.59274,0.54464)
		rgb=(0.12183,0.58905,0.54562)
		rgb=(0.12261,0.58537,0.54656)
		rgb=(0.12346,0.58169,0.54744)
		rgb=(0.12440,0.57800,0.54829)
		rgb=(0.12539,0.57432,0.54909)
		rgb=(0.12645,0.57063,0.54984)
		rgb=(0.12757,0.56695,0.55056)
		rgb=(0.12873,0.56327,0.55123)
		rgb=(0.12993,0.55958,0.55186)
		rgb=(0.13117,0.55590,0.55246)
		rgb=(0.13244,0.55222,0.55302)
		rgb=(0.13374,0.54853,0.55354)
		rgb=(0.13507,0.54485,0.55403)
		rgb=(0.13641,0.54117,0.55448)
		rgb=(0.13777,0.53749,0.55491)
		rgb=(0.13915,0.53381,0.55530)
		rgb=(0.14054,0.53013,0.55566)
		rgb=(0.14194,0.52645,0.55599)
		rgb=(0.14334,0.52277,0.55629)
		rgb=(0.14476,0.51909,0.55657)
		rgb=(0.14618,0.51541,0.55682)
		rgb=(0.14761,0.51173,0.55705)
		rgb=(0.14904,0.50805,0.55725)
		rgb=(0.15048,0.50437,0.55743)
		rgb=(0.15192,0.50069,0.55759)
		rgb=(0.15336,0.49700,0.55772)
		rgb=(0.15482,0.49331,0.55784)
		rgb=(0.15627,0.48962,0.55794)
		rgb=(0.15773,0.48593,0.55801)
		rgb=(0.15919,0.48224,0.55807)
		rgb=(0.16067,0.47854,0.55812)
		rgb=(0.16214,0.47484,0.55814)
		rgb=(0.16362,0.47113,0.55815)
		rgb=(0.16512,0.46742,0.55814)
		rgb=(0.16662,0.46371,0.55812)
		rgb=(0.16813,0.45999,0.55808)
		rgb=(0.16965,0.45626,0.55803)
		rgb=(0.17118,0.45253,0.55797)
		rgb=(0.17272,0.44879,0.55788)
		rgb=(0.17427,0.44504,0.55779)
		rgb=(0.17584,0.44129,0.55768)
		rgb=(0.17742,0.43753,0.55756)
		rgb=(0.17902,0.43376,0.55743)
		rgb=(0.18063,0.42997,0.55728)
		rgb=(0.18226,0.42618,0.55712)
		rgb=(0.18390,0.42238,0.55694)
		rgb=(0.18556,0.41857,0.55675)
		rgb=(0.18723,0.41475,0.55655)
		rgb=(0.18892,0.41091,0.55633)
		rgb=(0.19063,0.40706,0.55609)
		rgb=(0.19236,0.40320,0.55584)
		rgb=(0.19410,0.39932,0.55556)
		rgb=(0.19586,0.39543,0.55528)
		rgb=(0.19764,0.39153,0.55497)
		rgb=(0.19943,0.38761,0.55464)
		rgb=(0.20124,0.38367,0.55429)
		rgb=(0.20306,0.37972,0.55393)
		rgb=(0.20490,0.37575,0.55353)
		rgb=(0.20676,0.37176,0.55312)
		rgb=(0.20862,0.36775,0.55268)
		rgb=(0.21050,0.36373,0.55221)
		rgb=(0.21240,0.35968,0.55171)
		rgb=(0.21430,0.35562,0.55118)
		rgb=(0.21621,0.35153,0.55063)
		rgb=(0.21813,0.34743,0.55004)
		rgb=(0.22006,0.34331,0.54941)
		rgb=(0.22199,0.33916,0.54875)
		rgb=(0.22393,0.33499,0.54805)
		rgb=(0.22586,0.33081,0.54731)
		rgb=(0.22780,0.32659,0.54653)
		rgb=(0.22974,0.32236,0.54571)
		rgb=(0.23167,0.31811,0.54483)
		rgb=(0.23360,0.31383,0.54391)
		rgb=(0.23553,0.30953,0.54294)
		rgb=(0.23744,0.30520,0.54192)
		rgb=(0.23935,0.30085,0.54084)
		rgb=(0.24124,0.29648,0.53971)
		rgb=(0.24311,0.29209,0.53852)
		rgb=(0.24497,0.28768,0.53726)
		rgb=(0.24681,0.28324,0.53594)
		rgb=(0.24863,0.27877,0.53456)
		rgb=(0.25043,0.27429,0.53310)
		rgb=(0.25219,0.26978,0.53158)
		rgb=(0.25394,0.26525,0.52998)
		rgb=(0.25565,0.26070,0.52831)
		rgb=(0.25732,0.25613,0.52656)
		rgb=(0.25897,0.25154,0.52474)
		rgb=(0.26057,0.24692,0.52283)
		rgb=(0.26214,0.24229,0.52084)
		rgb=(0.26366,0.23763,0.51876)
		rgb=(0.26515,0.23296,0.51660)
		rgb=(0.26658,0.22826,0.51435)
		rgb=(0.26797,0.22355,0.51201)
		rgb=(0.26931,0.21882,0.50958)
		rgb=(0.27059,0.21407,0.50705)
		rgb=(0.27183,0.20930,0.50443)
		rgb=(0.27301,0.20452,0.50172)
		rgb=(0.27413,0.19972,0.49891)
		rgb=(0.27519,0.19490,0.49600)
		rgb=(0.27619,0.19007,0.49300)
		rgb=(0.27713,0.18523,0.48990)
		rgb=(0.27801,0.18037,0.48670)
		rgb=(0.27883,0.17549,0.48340)
		rgb=(0.27957,0.17060,0.48000)
		rgb=(0.28025,0.16569,0.47650)
		rgb=(0.28087,0.16077,0.47290)
		rgb=(0.28141,0.15583,0.46920)
		rgb=(0.28189,0.15088,0.46541)
		rgb=(0.28229,0.14591,0.46151)
		rgb=(0.28262,0.14093,0.45752)
		rgb=(0.28288,0.13592,0.45343)
		rgb=(0.28307,0.13090,0.44924)
		rgb=(0.28319,0.12585,0.44496)
		rgb=(0.28323,0.12078,0.44058)
		rgb=(0.28320,0.11568,0.43611)
		rgb=(0.28309,0.11055,0.43155)
		rgb=(0.28291,0.10539,0.42690)
		rgb=(0.28266,0.10020,0.42216)
		rgb=(0.28233,0.09495,0.41733)
		rgb=(0.28192,0.08967,0.41241)
		rgb=(0.28145,0.08432,0.40741)
		rgb=(0.28089,0.07891,0.40233)
		rgb=(0.28027,0.07342,0.39716)
		rgb=(0.27957,0.06784,0.39192)
		rgb=(0.27879,0.06214,0.38659)
		rgb=(0.27794,0.05632,0.38119)
		rgb=(0.27702,0.05034,0.37572)
		rgb=(0.27602,0.04417,0.37016)
		rgb=(0.27495,0.03775,0.36454)
		rgb=(0.27381,0.03150,0.35885)
		rgb=(0.27259,0.02556,0.35309)
		rgb=(0.27131,0.01994,0.34727)
		rgb=(0.26994,0.01463,0.34138)
		rgb=(0.26851,0.00961,0.33543)
		rgb=(0.26700,0.00487,0.32942)
	  }
}
\pgfplotscreateplotcyclelist{pastelcolors}{%
  solid, pastelPurple, mark=none\\%
  solid, pastelBlue, mark=none\\%
  solid, pastelGreen, mark=none\\%
  solid, pastelRed, mark=none\\%
  solid, pastelMagenta, mark=none\\%
  solid, pastelOrange, mark=none\\%
  solid, pastelSeaGreen, mark=none\\%
}

\usepackage[keeplastbox]{flushend}

\usepackage[backend=bibtex,style=ieee,mincitenames=1,maxcitenames=2,maxbibnames=20,url=false,isbn=false,doi=false]{biblatex}
\usetikzlibrary{positioning,arrows}
\usepgfplotslibrary{fillbetween}

\def\BibTeX{{\rm B\kern-.05em{\sc i\kern-.025em b}\kern-.08em
    T\kern-.1667em\lower.7ex\hbox{E}\kern-.125emX}}
\begin{document}

% \title{Verification of Image-based Neural Network Controllers Over an Approximate Input Space\\
% }

\title{Verification of Image-based Neural Network Controllers Using Generative Models \\
}

% \author{\IEEEauthorblockN{Sydney M. Katz}
% \IEEEauthorblockA{\textit{Aeronautics and Astronautics}, \textit{Stanford University}\\
% smkatz@stanford.edu}
% }

% \author{Sydney M. Katz, Anthony L. Corso, Christopher A. Strong, and Mykel J. Kochenderfer
% % <-this % stops a space
% %\thanks{*Denotes equal contribution}% <-this % stops a space
% \thanks{\textit{Stanford University}, Stanford, CA 94305, \{smkatz, acorso, castrong, mykel\}@stanford.edu}%
% }

\author{
    \IEEEauthorblockN{Sydney M. Katz\IEEEauthorrefmark{2}\IEEEauthorrefmark{1}, Anthony L. Corso\IEEEauthorrefmark{2}\IEEEauthorrefmark{1}, Christopher A. Strong\IEEEauthorrefmark{3}\IEEEauthorrefmark{1}, and Mykel J. Kochenderfer\IEEEauthorrefmark{2}}
    \IEEEauthorblockA{\IEEEauthorrefmark{2}Department of Aeronautics and Astronautics, Stanford University,
    \{smkatz, acorso, mykel\}@stanford.edu}
    \IEEEauthorblockA{\IEEEauthorrefmark{3}Department of Electrical Engineering, Stanford University,
    castrong@stanford.edu}
    \IEEEauthorblockA{\IEEEauthorrefmark{1}Denotes equal contribution}
    %\thanks{*Denotes equal contribution}
    }

\maketitle

\begin{abstract}
Neural networks are often used to process information from image-based sensors to produce control actions. While they are effective for this task, the complex nature of neural networks makes their output difficult to verify and predict, limiting their use in safety-critical systems. For this reason, recent work has focused on combining techniques in formal methods and reachability analysis to obtain guarantees on the closed-loop performance of neural network controllers. However, these techniques do not scale to the high-dimensional and complicated input space of image-based neural network controllers. 
% However, these techniques do not scale to the high-dimensional and poorly defined input space of image-based neural network controllers.
In this work, we propose a method to address these challenges by training a generative adversarial network (GAN) to map states to plausible input images.
% In this work, we propose a method to address these challenges by training a generative adversarial network (GAN) to approximate the input space.
By concatenating the generator network with the control network, we obtain a network with a low-dimensional input space. This insight allows us to use existing closed-loop verification tools to obtain formal guarantees on the performance of image-based controllers. We apply our approach to provide safety guarantees for an image-based neural network controller for an autonomous aircraft taxi problem. We guarantee that the controller will keep the aircraft on the runway and guide the aircraft towards the center of the runway. The guarantees we provide are with respect to the set of input images modeled by our generator network, so we provide a recall metric to evaluate how well the generator captures the space of plausible images.
%The guarantees we provide are with respect to the input space modeled by our generator network, so we provide a recall metric to evaluate how well the generator captures the true input space.
\end{abstract}

\section{Introduction} \label{sec:intro}
Due to their ability to represent complex functions, neural networks are well-suited to perform control tasks in complicated and high-dimensional problem settings \cite{mnih2015human, pan2017agile}. Proposed autonomous aviation systems, for instance, use neural networks to perform safety critical tasks such as collision avoidance, autonomous taxiing, and autonomous landing \cite{Julian2016dasc, julian2020validation, byun2020manifold, cofer2020run}. In some instances, these systems process images using deep neural networks that are trained to produce safe % and effective 
control actions. For example, recent work has shown that a neural network controller that takes in images from a camera placed on the wing of an aircraft can effectively guide it down the center of a runway \cite{julian2020validation, byun2020manifold, cofer2020run}. However, to use neural networks in safety-critical applications, we must develop techniques to verify that they will operate safely. 
Although effective for their proposed applications, image-based neural network controllers are difficult to verify due to the high-dimensional and complicated input space, the complexity of deep neural networks, and the closed-loop nature of the control problem.
%Although effective for their proposed applications, image-based neural network controllers are difficult to verify due to the large and poorly characterized input space, the complexity of deep neural networks, and the closed-loop nature of the control problem.

Recent work in formal methods has resulted in tools that are able to verify input-output properties of neural networks \cite{reluval, Katz2017, katz2019marabou, tjeng2017evaluating, liu2019algorithms, gehr2018ai2}. Neural network verification tools take as input a bounded region in the input space of the network and provide guarantees on properties of the output space \cite{liu2019algorithms}. For example, for a network that outputs control actions, a neural network verification tool can provide guarantees on the actions the agent may take in a given region of the state space. Existing closed-loop verification techniques combine the output of these tools with techniques in reachability analysis to provide guarantees on the closed-loop performance of neural network controllers \cite{katz2021probabilistic, Julian2019dasc, huang2019reachnn, xiang2019reachable, Sidrane2019iclr}. However, while these approaches work well for state-based controllers with low-dimensional input spaces, they do not scale to image-based controllers.
%However, while these approaches work well for state-based controllers with well-defined, low-dimensional input spaces, they do not scale to image-based controllers.

% For the autonomous taxiing problem, a state-based controller requires a two-dimensional input consisting of the aircraft's crosstrack and heading error, while an image-based controller requires a 128-dimensional image input. Furthermore, the input space of an image-based controller is not well-defined. While we can easily specify a range of values that we expect each state variable to take on, bounding the input space of an image-based controller is less straightforward. To bound the input space for the aircraft taxi problem, we would need to characterize all images in pixel space that look like a runway. Therefore, image-based controllers present two major verification challenges over state-based controllers: the input space is high-dimensional and not well-defined.

For the autonomous taxiing problem, a state-based controller requires a two-dimensional input consisting of the aircraft's crosstrack and heading error, while an image-based controller requires a 128-dimensional image input. Furthermore, the safety properties we would like to verify are more naturally expressed in the state space rather than the image space. For instance, to perform closed-loop verification, we need to select the region of the input space for which we expect our safety properties to hold. While we can easily specify a range of values that we expect each state variable to take on, selecting a region in the input space of an image-based controller is less straightforward. To specify the set of plausible input images for the aircraft taxi problem, we would need to define all images in pixel space that look like a runway. 
%Therefore, image-based controllers present two major verification challenges over state-based controllers: the input space is high-dimensional and not well-defined.

Because of these challenges, the verification community has mainly focused on the problem of adversarial robustness, where we find the maximum perturbation of the output for a bounded noise disturbance \cite{Strong2020, chakraborty2018adversarial, carlini2017provably}. This approach was used to find sequences of noise disturbances that could cause a neural network to guide the aircraft off the runway \cite{julian2020validation}. However, this approach lacks guarantees and is only able to capture image perturbations due to noise, missing out on other possible semantic variations such as lighting conditions and skid marks. 

Generative models are well-suited to capture this semantic variation, and other work has performed verification across line segments in the latent space of a generative adversarial network (GAN) to determine the robustness of a classifier to changes in orientation for human faces \cite{mirman2020robustness}. Furthermore, GANs have been used to find erroneous behavior of neural networks in driving scenarios \cite{zhang2018deeproad, hanapply}. However, these works focus on using a GAN to find failures at a single point in time rather than reasoning about the entire closed-loop system. 
%Additionally, they do not take advantage of neural network verification tools to guide their search. 
While techniques exist to verify the closed-loop performance of perception-based controllers, they rely on the specification of an exact geometric mapping between the current state and the input to the controller \cite{sun2019formal, yang2019correctness}. Such a mapping often does not exist in complicated, real-world scenarios.

% Generative models are well-suited to capture this semantic variation, and other work has made use of generative adversarial network (GANs) to find erroneous behavior of neural networks in driving scenarios \cite{zhang2018deeproad, hanapply}. They focus, however, on using a GAN to find failures at a single point in time rather than reasoning about the entire closed-loop system. 
% Additionally, they do not take advantage of neural network verification tools to guide their search. Other work performs verification across line segments in the latent space of a generative model to determine the robustness of a classifier for human faces \cite{mirman2020robustness}.

In this work, we present a technique to characterize the set of plausible inputs for an image-based neural network controller that addresses the challenges of verifying image-based controllers. Our technique relies on training a conditional GAN (cGAN) to produce realistic input images for a given state and concatenating the generator network with the control network. In doing so, we obtain a network that goes from a low-dimensional state-based input to a control output rather than from a high-dimensional image input to a control output. This method effectively reduces the verification problem to that of verifying a state-based neural network controller for which we can rely on existing techniques \cite{katz2021probabilistic, Julian2019dasc, huang2019reachnn, xiang2019reachable, Sidrane2019iclr}.

We use our approach to verify the closed-loop performance of an image-based neural network controller for the autonomous taxi problem. We provide guarantees that the controller will keep the aircraft on the runway and that the aircraft trajectory will converge to an area near the center of the runway within a few seconds for the set of all images that could be produced by the generator network. Finally, because the safety guarantees we provide are only as strong as the expressiveness of the generative model, we provide a recall metric to quantify how well the generator network from the GAN captures the space of plausible images.

\section{Background}\label{sec:background}
This work builds upon previously developed techniques for verifying state-based neural network controllers and creating generative models. This section details the necessary background on these topics.

\subsection{Neural Network Verification}\label{sec:nnv}
Neural network verification tools formally reason about input-output properties of neural networks \cite{liu2019algorithms}. They answer yes or no questions %\todo{not sure I like the wording of binary guarantees? better way to say this? EDIT: updated to be yes or no (Chris)} 
about whether a certain input-output property holds. Specifically, for a neural network representing the function $y = f_{n}(x)$, these tools can determine whether the property
\begin{equation}
\label{eq:verification_problem}
    x \in \mathcal{X} \implies f_{n}(x) \notin \mathcal{Y}
\end{equation}
holds for convex polytopes $\mathcal{X}$ and $\mathcal{Y}$. Recent work has shown that these tools can be extended to an optimization framework, in which we solve the following optimization problem
\begin{equation} \label{eq:output_opt}
        \begin{aligned}
            & \underset{x}{\text{minimize}} && g(f_{n}(x)) \\
            & \text{subject to} && x \in \mathcal{X}
        \end{aligned}
\end{equation}
where $g$ is a convex function \cite{Strong2020}.

In the context of state-based neural network controllers, the input to the network is the state of the system, and the output is the corresponding control action. Thus, we can use neural network verification tools to determine the range of actions possible in a given region of the state space. Previous work on problems with discrete action spaces used neural network verification tools to determine whether a particular action is possible in a given region \cite{Julian2019dasc, katz2021probabilistic}. To extend to problems with continuous action spaces, we use the optimization-based framework to determine the minimum and maximum control output in a given region of the state space. 

%%%%% This figure goes with the approach section but is here to fix figure placement
\begin{figure*}[t!]
    \centering
    \input{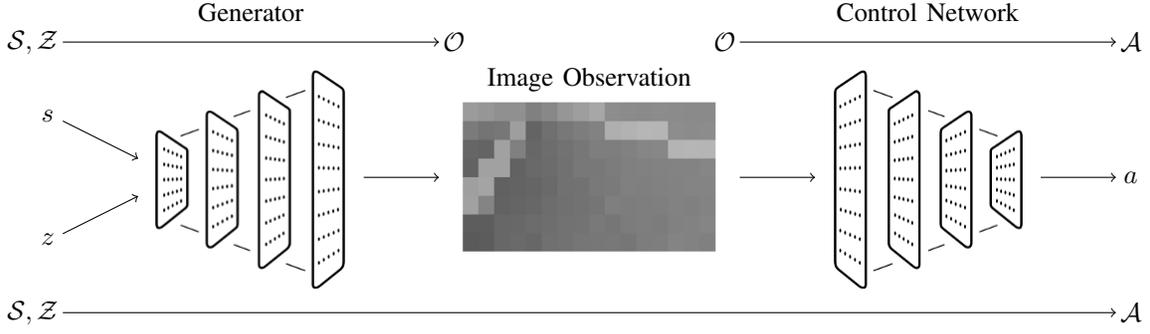}
    \caption{Overview of approach to verification for image-based controllers. The generator and control network are concatenated to obtain a network that goes from a low-dimensional, well-defined space to a control action. The concatenated network can be used with existing closed-loop verification techniques. \label{fig:approach_overview}}
\end{figure*}

\subsection{Verification of State-based Neural Network Controllers}\label{sec:state_based_verif}
%While we can use neural network verification tools to verify input-output properties that should hold based on intuition, this process does not guarantee safety. 
To provide guarantees on safety properties of neural network controllers, we must evaluate their closed-loop performance. A number of techniques exist to verify the closed-loop performance of state-based neural network controllers \cite{katz2021probabilistic, Julian2019dasc, huang2019reachnn, xiang2019reachable, Sidrane2019iclr}. In general, these methods combine the output of neural network verification tools with techniques in reachability analysis to reason about closed-loop properties. In this section, we formalize two methodologies that we use in this work to verify properties \cite{Julian2019dasc, katz2021probabilistic}.

We assume that we are given a neural network controller that represents a policy $\pi$, which is a function that maps states in a bounded state space $\mathcal{S}$ to an action in the action space $\mathcal{A}$. Additionally, we assume that we are provided with a dynamics model $s_{t+1}= f(s_t, a_t)$ that maps state-action pairs to their corresponding next state. In order to make verification tractable over the entire state space $\mathcal{S}$, we divide the state space into a finite number of hyperrectangular cells $c \in \mathcal{C}$. We then use a neural network verification tool as described in \cref{sec:nnv} to determine the set of actions possible in a cell, which we denote as $\mathcal{A}_c$. Using these specifications, we define an overapproximated dynamics model $f(c, \mathcal{A}_c)$ that maps cells and their corresponding action ranges to a set of reachable next cells $\mathcal{C}^\prime$.

One safety property of interest is whether we can reach a set of unsafe states $\mathcal{B}$. In the aircraft taxi problem, for example, we are interested in determining whether we could leave the runway. Let $F^\pi(c) \in \{0, 1\}$ represent the possibility of reaching a cell in $\mathcal{B}$ given that we start in cell $c$ and follow neural network policy $\pi$. This quantity can be determined according to the methods outlined in \citeauthor{katz2021probabilistic} \cite{katz2021probabilistic}. For all cells $c \in \mathcal{B}$, $F^\pi(c) = 1$. For all cells $c \notin \mathcal{B}$, $F^\pi(c)$ is written recursively as  
\begin{equation}
    F^\pi(c) = \max_{c' \in f(c, \mathcal{A}_c)} F^\pi(c')
\end{equation}
and can be computed using dynamic programming. Because we use an overapproximated dynamics model, the results of the analysis will also represent an overapproximation. Thus, for cells where $F^\pi(c) = 1$, the analysis is inconclusive in that an unsafe state may be reachable, while cells with $F^\pi(c) = 0$ are guaranteed to be safe.

Given a set of initial states, another property of interest is the forward reachable set, which may allow us to determine if we reach a set of goal states. For instance, in the aircraft taxi problem, we want to reach a set of states near the center of the runway. For this task, we follow the methodology defined by \citeauthor{Julian2019dasc} to find an overapproximated reachable set \cite{Julian2019dasc}. Let the function $\text{succ}(c)$ return the set of cells from which cell $c$ can be reached according to our overapproximated dynamics model when following policy $\pi$. We perform forward reachability as follows 
\begin{equation}
    R^\pi_{t+1}(c) = \max_{c' \in \text{succ}(c)} R^\pi_{t}(c')
\end{equation}
for time $t>0$, where $R^\pi_{t}(c)$ represents the possibility of reaching cell $c$ at time step $t$ when following policy $\pi$. At $t=0$, $R^\pi_0(c) = 1$ for all cells that overlap with the initial state region and $R^\pi_0(c) = 0$ otherwise.
%At $t=0$, all cells are set to a value of one if they overlap with the initial state region and zero otherwise.

\subsection{Generative Adversarial Networks}
\label{sec:GANs}
Generative adversarial networks ~\cite{goodfellow2014generative, creswell2018generative} are a class of machine learning models used to generate samples that are similar to a training dataset. A GAN consists of a generator and discriminator engaged in a zero-sum game where the discriminator is trained to distinguish real and generated data, while the generator is trained to fool the discriminator. GANs have been used to generate realistic looking images of faces and commonplace objects~\cite{karras2019style}, to augment image-based datsets for medical applications~\cite{frid2018gan}, and to transfer artistic styles between images~\cite{zhu2017unpaired}. 

In this work, we focus on cGANs where the training data $X$ with associated label $Y$ comes from the conditional distribution $P_r(X \mid Y)$. The generator model $G$ induces a distribution $P_g(X \mid Y)$ by mapping a randomly distributed latent vector $Z \sim P_Z$ and label $Y \sim P_Y$ to a sample $X = G(Z,Y)$. The discriminator model $D(X,Y)$ takes in a sample $X$ and its corresponding label $Y$ and outputs the probability that the sample came from the true data distribution, $P_r$, rather than $P_g$. The generator is trained to fool the discriminator by minimizing the loss 
\begin{equation}
    \mathcal{L_G} = \mathbb{E}_{Y\sim P_Y, Z\sim P_Z}  \left[ \log (1 - D(G(Z,Y), Y)) \right]
\end{equation}
% \begin{equation}
%     \mathcal{L_G} = \frac{1}{N} \sum_{z_i, y_i} \left[ \log (1 - D(G(z_i, y_i), y_i)) \right]
% \end{equation}
while the discriminator is trained by minimizing the loss
\begin{equation}
\begin{split}
    \mathcal{L}_D = &\mathbb{E}_{X \sim P_r(X \mid Y), Y\sim P_Y}\left[ \log (1-D(X, Y)) \right] + \\ &\mathbb{E}_{Y\sim P_Y, Z\sim P_Z} \left[ \log D(G(Z,Y), Y) \right]
\end{split}
\end{equation}
to distinguish real and generated data.

\section{Approach}
% \begin{figure*}[t!]
%     \centering
%     \input{approach_overview}
%     \caption{Overview of approach to verification for image-based controllers. The generator and control network are concatenated to obtain a network that goes from a low-dimensional, well-defined space to a control action. The concatenated network can be used with existing closed-loop verification techniques. \label{fig:approach_overview}}
% \end{figure*}

An image-based neural network controller uses image observations of the state to determine the control action. We define the region of the input space for which we expect our safety properties to hold as the observation space $\mathcal{O}$, where each observation corresponds to a set of values (RGB or grayscale) for each pixel in the image. For this type of neural network controller, we cannot directly apply the state-based verification methods described in \cref{sec:state_based_verif}. Because the number of cells required by these methods scales exponentially with the input dimension, the high-dimensional input space of image-based controllers makes the problem intractable. Furthermore, while we can easily specify bounds on the region of the input space we would like to verify over for state-based controllers, defining this space for image-based controllers tends to be less straightforward. To define the observation space for an image-based controller, we must quantify the set of images in pixel space that we expect our controller to encounter. 

% An image-based neural network controller uses image observations of the state to determine the control action. We define its input space as a set of observations $o \in \mathcal{O}$, where each observation corresponds to a set of values (RGB or grayscale) for each pixel in the image. For this type of neural network controller, we cannot directly apply the state-based verification methods described in \cref{sec:state_based_verif}. Because the number of cells required by these methods scales exponentially with the input dimension, the high-dimensional input space of image-based controllers makes the problem intractable. Furthermore, while we can easily specify bounds on the region of the state space we would like to verify over for state-based controllers, defining the set of plausible inputs for image-based controllers tends to be less straightforward. To specify the space we would like to verify over for an image-based controller, we must quantify the set of images in pixel space that we expect our controller to encounter. 

\subsection{Observation Space Approximation}
%While quantifying the exact input space of an image-based controller is impractical \todo{intractable? impractical? both?}, we propose to approximate the space by training a GAN on a representative set of samples from the observation space. 
We propose to approximate the observation space of an image-based controller by training a GAN on a representative set of samples. Our overall approach is described in \cref{fig:approach_overview}. We train a generator network using the GAN framework to map states in a bounded state space $\mathcal{S}$ and latent variables in a bounded latent space $\mathcal{Z}$ to image observations in $\mathcal{O}$. The controller network we would like to verify takes in observations from $\mathcal{O}$ and outputs an action in $\mathcal{A}$. Noting that the output space of the generator network corresponds to the input space of the control network, we concatenate the two networks to obtain a network that maps from states and latent variables to control actions. The result is a network that maps from a low-dimensional, well-defined space of states and latent variables to control actions, which we can verify using the techniques outlined in \cref{sec:state_based_verif}.

% While quantifying the exact input space of an image-based controller is impractical \todo{intractable? impractical? both?}, we propose to approximate the space by training a GAN on a representative set of samples from the observation space. Our overall approach is described in \cref{fig:approach_overview}. We train a generator network $G(s, z)$ using the GAN framework to model $P(O \mid S)$. The resulting network maps states in a bounded state space $\mathcal{S}$ and latent variables in a bounded latent space $\mathcal{Z}$ to image observations in $\mathcal{O}$. The controller network $C(o)$ we would like to verify takes in observations from $\mathcal{O}$ and outputs an action in $\mathcal{A}$. Noting that the output space of the generator network corresponds to the input space of the control network, we concatenate the two networks to obtain a network that maps from states and latent variables to control actions. The result is a network that maps from a low-dimensional, well-defined space of states and latent variables to control actions, which we can verify using the techniques outlined in \cref{sec:state_based_verif}.

\subsection{Evaluation Metric}
The properties we verify using our proposed approach are guaranteed to hold with respect to the observation space approximated by the generator network from the GAN. As a result, to make statements about the overall system, we must quantify how well the generator approximates the true observation space. In particular, we are interested in the recall of our generator network, which represents the probability that a random image sampled from the training data falls under the image of the generator network \cite{kynkaanniemi2019improved}. Assuming that our training data provides a representative sample of the observation space, this metric evaluates the quality of the generator's approximation.
% In particular, we are interested in the recall of our generator network. Assuming that our training data provides a representative sample of the true observation space, the recall represents the probability that a random image sampled from the training data falls under the image of the generator network \cite{kynkaanniemi2019improved}.
% represented by the training data. In particular, we are interested in the recall of our generator network, which represents the probability that a random image sampled from the training data falls under the image of the generator network \cite{kynkaanniemi2019improved}. 
However, because images are made up of high-dimensional floating point values, it is unlikely that our generator will produce the training images exactly. Therefore, we instead analyze the distance between each training point and its nearest neighbor in the generator space according to their Euclidean distance.

Let the generator network represent the function $o = G(z, s)$. For a given training image $o_i$, we want to find the distance to the closest image in the generator output space as follows
\begin{equation} \label{eq:generator_opt}
        \begin{aligned}
            & \underset{z}{\text{minimize}} && \|G(z, s_i) - o_i\|_2 \\
            & \text{subject to} && z \in \mathcal{Z}
        \end{aligned}
\end{equation}
where $s_i$ corresponds to the state represented by $o_i$. This problem matches the form of \cref{eq:output_opt}, so we can solve it using an optimization-based neural network verification framework. We solve this optimization problem for each point in the training data to obtain a set of distances for a particular generator network. By taking the cumulative distribution of these distances, we obtain a notion of recall that we can use to compare the effectiveness of various generator networks.

\section{Application: Aircraft Taxi Problem}
We demonstrate our approach on the autonomous aircraft taxi problem \footnote{Source is at \href{https://github.com/sisl/VerifyGAN}{https://github.com/sisl/VerifyGAN}.}, which has recently been used as a benchmark problem in work on robust and verified perception \cite{julian2020validation, byun2020manifold}. In previous work, a neural network was trained to take images from a camera on the right wing of a Cessna 208B Grand Caravan taxiing at 5 m/s down runway 04 of Grant County International Airport and output a control action (steering angle) that keeps the aircraft on the runway \cite{julian2020validation}. In this work, we aim to verify that an aircraft using this controller will abide by the following two safety properties:

\begin{itemize}
    \item Property 1 (\textbf{P1}): The aircraft will not leave the runway.
    \item Property 2 (\textbf{P2}): The aircraft will be guided towards the center of the runway.
\end{itemize}

The state of the aircraft is characterized by its crosstrack position $p$ and heading error $\theta$. We write the system dynamics $f([p, \theta], \phi)$ as follows
\begin{equation}
    \begin{split}
        p & \leftarrow p + v \Delta t \sin \theta \\
        \theta & \leftarrow \theta + \frac{v}{L}\Delta t \tan \phi
    \end{split}
\end{equation}
where $\Delta t$ is the time step, $v$ is the taxi speed (\SI{5}{\meter\per\second}), $L$ is the distance between the front and back wheels (\SI{5}{\meter}), and $\phi$ is the steering angle control input. The neural network controller outputs predictions for $p$ and $\theta$, which can be used to determine the steering angle $\phi$ using the following proportional control law
\begin{equation}\label{eq:prop_control}
    \phi = -0.74p - 0.44\theta
\end{equation}
The image observations are obtained using the X-Plane 11 flight simulator \cite{xplane11}.

\subsection{GAN Training}
\begin{figure}[t!]
    \centering
    \input{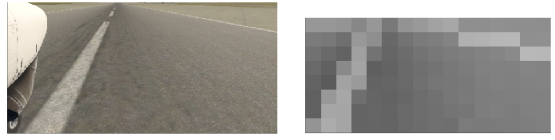}
    \caption{Example image of the runway from a camera on the wing of the aircraft (left) and corresponding downsampled image (right). \label{fig:im_ex}}
\end{figure}
\begin{figure*}[t!]
    \centering
    \input{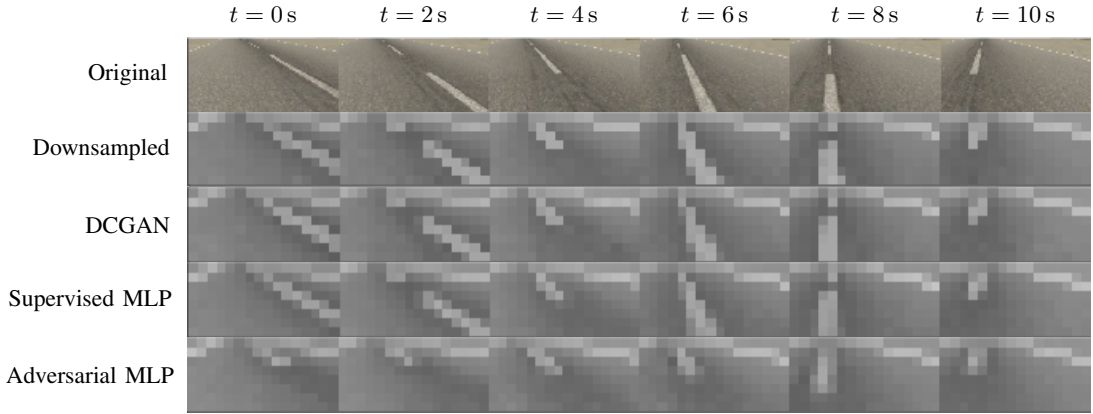}
    \caption{True images, downsampled images, and closest generated images at various points in the aircraft trajectory. While the DGCAN and Supervised MLP closely match the true images, the Adversarial MLP shows signs of mode collapse. \label{fig:gen_comp}}
\end{figure*}
We obtained the training data used in this work from a \num{200}-meter portion of the runway that starts approximately \num{300} meters down the runway. Images were obtained for \num{10000} uniformly sampled states with crosstrack position ranging from \num{-11} to \num{11} meters and heading error ranging from \num{-30} to \num{30} degrees. To match the input required by the neural network controller, each image is cropped, converted to grayscale, downsampled to increase the brightness of the runway markings, and biased so that all pixels have an average value of \num{0.5} \cite{julian2020validation}. \Cref{fig:im_ex} shows an example runway image alongside the corresponding downsampled image. The result is a dataset of \num{10000} downsampled $8 \times 16$ images that represent a sampling from the space of images that may be input to the neural network controller.

Deep convolutional GANs (DCGANs) have been shown to perform well in image generation tasks \cite{radford2015unsupervised}. However, most neural network verification tools require networks in the form of multilayer perceptrons (MLPs) with piecewise linear activation functions such as rectified linear unit (ReLU) activations \cite{liu2019algorithms}. Therefore, training a generator that meets the requirements of existing neural network verification tools while still producing realistic images requires specific design considerations. We accomplish this by first training a traditional DCGAN and subsequently using ideas from GAN distillation to train a smaller MLP network \cite{li2020gan}.

The generator takes as input the state of the aircraft along with a two-dimensional latent vector in which each entry is sampled independently from a uniform distribution between $-1$ and $1$. Because the state only contains the crosstrack position and heading error of the aircraft, the latent variables are meant to capture semantic variations caused by changes in downtrack position such as variation in the visibility of the centerline dash. For the DCGAN, the network architectures of the generator and discriminator were those used by \citeauthor{miyato2018spectral} with a small modification to condition on the state \cite{miyato2018spectral}. Specifically, we used one dense layer in the generator to map the latent variables to a $2 \times 1 \times 16$ tensor and another dense layer to map the state to a $2 \times 1 \times 496$ tensor. Both tensors were concatenated to construct a $2 \times 1 \times 512$ tensor as the input to the generator. The discriminator included an extra dense layer that mapped the state to an $8 \times 16$ matrix which was concatenated to the input image to create an $8 \times 16 \times 2$ input.

The training was performed using the loss functions described in \cref{sec:GANs}. To stabilize the training, we applied several normalization and regularization techniques. Spectral normalization was applied to all layers in the discriminator~\cite{miyato2018spectral}. We initialized all layers in both models using orthogonal initialization and applied orthogonal regularization to the generator loss function~\cite{brock2017neural}. The training hyperparameters are shown in \cref{tab:hyperparams}. 

\begin{table}
    \centering
    \caption{Hyperparameters for GAN Training \label{tab:hyperparams}}
    \begin{tabular}{@{}ll@{}} 
        \toprule
        \textbf{Parameter} & \textbf{Value} \\
        \midrule
        Real examples & \num{10000} \\
        Batch size & \num{256} \\
        Learning rate & \num{7e-4} \\
        Epochs & \num{750} \\
        Optimizer & ADAM($\beta=(0.5, 0.99)$) \\
        Orthogonal regularization parameter & \num{1e-4} \\
        \bottomrule
    \end{tabular}
\end{table}

Because the discriminator is not used in the verification portion of this work, we only need to simplify the generator to a ReLU-activated MLP. We explored two simplification approaches. The first was to train an MLP generator $G_{\rm MLP}$ as a GAN with the pretrained DCGAN discriminator. 
%We found, however that the generator suffered from mode-collapse~\cite{gulrajani2017improved} and therefore poorly represented the space of images, as shown in the last row of \cref{fig:im_comp}. 
The second approach relied on supervised training of the MLP with data produced by the DCGAN generator $G_{\rm DCGAN}$. In particular, we trained the MLP to minimize the loss
\begin{equation}
\begin{split}
    \frac{1}{N} \sum_{i=1}^N &\| G_{\rm DCGAN}(z_i, s_i) - G_{\rm MLP}(z_i, s_i) \|_p - \\ &\lambda D(G_{\rm MLP}(z_i, s_i), s_i)
\end{split}
\end{equation}
where $z_i$ and $s_i$ are sampled uniformly in the latent space and state space, respectively. We included the discriminator to improve the quality of the generated samples as suggested by \citeauthor{pathak2016context} \cite{pathak2016context}. A coarse search of hyperparameters found that the best MLP images (as measured by mean squared error to the DCGAN images) were produced with $p=1$, $\lambda=\num{7e-3}$, and a learning rate of \num{1e-3}. The MLP architecture was chosen to have \num{4} hidden layers of \num{256} units, as this was large enough to produce realistic images but small enough to be used with the neural network verification tool. 
%\todo{should we note somewhere that we used a uniform distribution? Anthony add this and Sydney add truncation verification}

% \begin{figure*}[t]
%     \centering
%     \input{generator_comp}
%     \caption{True images, downsampled images, and closest generated images at various points in the aircraft trajectory. While the DGCAN and Supervised MLP closely match the true images, the Adversarial MLP shows signs of mode collapse. \label{fig:gen_comp}}
% \end{figure*}
\Cref{fig:gen_comp} shows a set of true images along with the closest (in Euclidean distance) generator images for each training architecture. Based on visual inspection, the DCGAN and supervised MLP generated images closely match the true images. The images generated by the MLP network trained using adversarial training, on the other hand, do not match as well. In particular, the network appears to only produce images between the dash marks on the runway. This mode collapse is a common issue that occurs in GAN training when the generator focuses on a small subset of images that fool the discriminator rather than learning the full distribution of training images. To supplement the visual comparison in \cref{fig:gen_comp}, we provide a quantitative evaluation of the performance of each generator in \cref{sec:recall_res}. Based on these results, we used the supervised MLP network for the rest of our analysis.

\subsection{Simulation}
\begin{figure}[htb]
    \centering
    \input{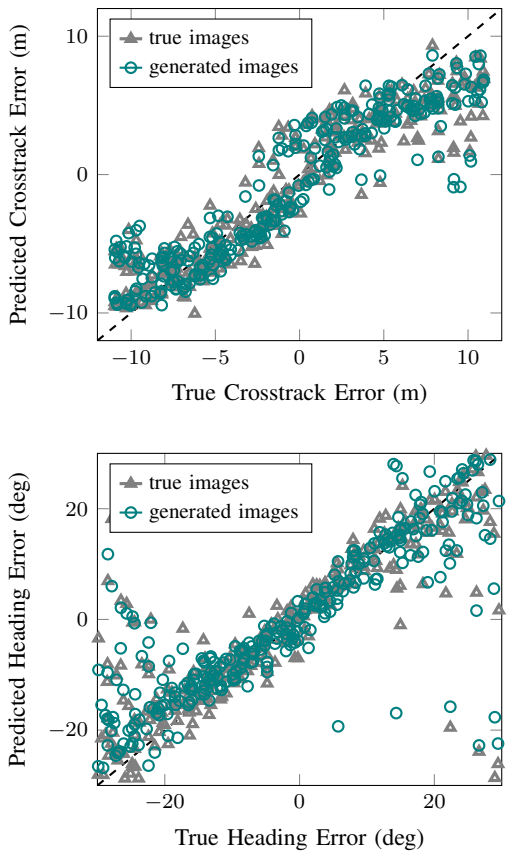}
    \caption{Comparison of the state predictions for the control network over a sample of true and generated images. The dashed line represents the line $y=x$. \label{fig:preds_comp}}
\end{figure}
\begin{figure}[htb]
    \centering
    \input{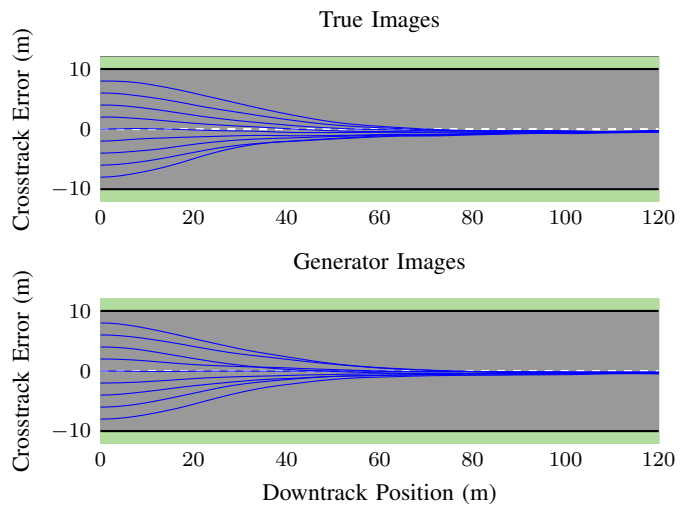}
    \caption{Taxi trajectories from various starting crosstrack positions on the runway with the true images as input (top) and generated images as input (bottom). \label{fig:sims}}
\end{figure}
Before applying formal verification techniques to the neural network controller, we can evaluate its prediction accuracy on a sampling of images. \Cref{fig:preds_comp} compares the network's prediction of crosstrack and heading error on both the generated and true images. The control network noisily tracks the true errors for both sets of input images. Additionally, the noise distribution of the predictions is similar across the generated and true images.

To further test the performance of the neural network controller, we can simulate its trajectory on the runway starting from various crosstrack positions using both the true and the generated images as input. \Cref{fig:sims} shows a comparison of the taxi trajectories of the aircraft from nine starting crosstrack positions. The trajectories appear to be similar across both input types. All trajectories satisfy the safety properties \textbf{P1} and \textbf{P2}, staying on the runway and converging to a location near the center of the runway. While these simulations support the hypothesis that the neural network controller satisfies the safety properties, they do not evaluate all possible trajectories and therefore do not provide a guarantee.

\subsection{Closed-loop Verification}
To provide guarantees on our desired safety properties, we use the closed-loop verification techniques described in \cref{sec:state_based_verif}. We divide the input space into \num{16384} uniform cells by splitting each dimension into \num{128} bins of equal width. Let $h(s, z)$ represent the function modeled by the concatenated generator and control network, which takes as input a state and latent variable instantiation and outputs the state estimate from the control network. Using a neural network verification tool, we solve the following optimization problem to determine the minimum control output for a given cell
\begin{equation} \label{eq:cell_opt}
        \begin{aligned}
            & \underset{x}{\text{minimize}} && [-0.74, -0.44]^\top h(s, z) \\
            & \text{subject to} && s \in c, z \in \mathcal{Z}
        \end{aligned}
\end{equation}
where the objective represents the proportional control law from \cref{eq:prop_control}.

Optimizing the negative version of the objective function provides us with the maximum control output, and together these results define $\mathcal{A}_c$. To perform the optimization, we use a modified version of the \textsc{DeepZ} verification algorithm \cite{singh2018fast}. We frame the optimization problem as a branch and bound search \cite{lawler1966branch,kochenderfer2019algorithms}, eagerly splitting the input space and applying \textsc{DeepZ} at each step to refine the bounds on the optimum. The true optimum can be found to a desired tolerance, which we set to \num{1e-4}. Analyzing our initial verification results revealed that the generator often produces images with undesirable artifacts for inputs on the edge of the latent space. As a result, we truncate the latent space to contain only inputs with values between \num{-0.8} and \num{0.8} rather than the full \num{-1} to \num{1} range. This truncation did not significantly degrade the performance of the generator according to our recall metric.

\begin{figure}[t!]
    \centering
    \input{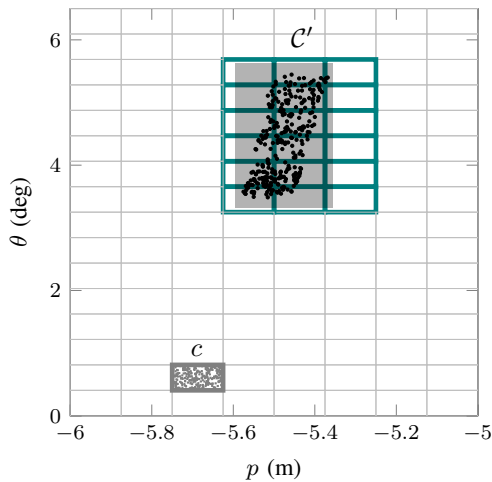}
    \caption{Example of overapproximated dynamics model. The gray and black dots represent sampled positions at time step $t$ and $t+1$ respectively when transitioning from cell $c$. The shaded region represents the result of applying \cref{eq:overapprox_dyn}, and the teal cells comprise $\mathcal{C}^\prime$. \label{fig:reach_demo}}
\end{figure}
Next, we define our overapproximated dynamics function $f(c, \mathcal{A}_c)$. Each cell $c$ defines a region of the input space where $p \in [p_{\min}, p_{\max}]$ and $\theta \in [\theta_{\min}, \theta_{\max}]$. Noting that the sine and tangent functions are monotonically increasing in our operating region, we define the overapproximated dynamics as follows
\begin{equation}\label{eq:overapprox_dyn}
    \begin{split}
        p & \leq  p_{\max} + v \Delta t \sin \theta_{\max} \\
        p & \geq  p_{\min} + v \Delta t \sin \theta_{\min} \\
        \theta & \leq \theta_{\max} + \frac{v}{L} \Delta t \tan \phi_{\max} \\
        \theta & \geq \theta_{\min} + \frac{v}{L} \Delta t \tan \phi_{\min} \\
    \end{split}
\end{equation}
where $\phi_{\min}$ and $\phi_{\max}$ represent the minimum and maximum steering angle according to $\mathcal{A}_c$. The set of next cells $\mathcal{C}^\prime$ is then defined as the set of cells in $\mathcal{C}$ that overlap with this region. \Cref{fig:reach_demo} shows this calculation for an example cell. We sampled $300$ points uniformly within the cell (gray) and propagated them through the dynamics (black). The shaded region represents the result of applying \cref{eq:overapprox_dyn}, and the highlighted cells represent the set of overlapping cells $\mathcal{C}^\prime$.
% \begin{figure}[htb]
%     \centering
%     \input{reachability_demo}
%     \caption{Example of overapproximated dynamics model. The gray and black dots represent sampled positions at time step $t$ and $t+1$ respectively when transitioning from cell $c$. The shaded region represents the result of applying \cref{eq:overapprox_dyn}, and the teal cells comprise $\mathcal{C}^\prime$. \label{fig:reach_demo}}
% \end{figure}

\begin{figure}[t!]
    \centering
    \begin{tikzpicture}[]
\begin{axis}[
  height = {7cm},
  ylabel = {$\theta$ (deg)},
  xmin = {-11.0},
  xmax = {11.0},
  ymax = {30.0},
  xlabel = {$p$ (m)},
  ymin = {-30.0},
  width = {7cm},
  enlargelimits = false,
  axis on top
]

\addplot[
  ] graphics[
  xmin = -11.0,
  xmax = 11.0,
  ymin = -30.0,
  ymax = 30.0
] {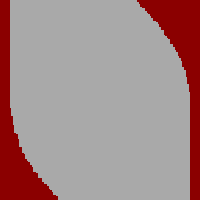};

\end{axis}
\end{tikzpicture}
    \caption{Set of states guaranteed to satisfy \textbf{P1} with respect to the input space defined by the generator network. Gray states are guaranteed to be safe, while red states are inconclusive. \label{fig:safe_cells}}
\end{figure}

With these definitions in place, we can apply the techniques described in \cref{sec:state_based_verif}. \Cref{fig:safe_cells} shows the result of applying the methods in \citeauthor{katz2021probabilistic} to test \textbf{P1} \cite{katz2021probabilistic}. If the aircraft starts in a gray cell, the neural network controller is guaranteed to keep the aircraft on the runway. If the aircraft starts in a red cell, whether the aircraft will leave the runway is inconclusive.
%there is a possibility that the neural network controller will guide the aircraft off the runway.
% \begin{figure}[t!]
%     \centering
%     \input{safe_cells}
%     \caption{Set of states guaranteed to satisfy \textbf{P1} with respect to the input space defined by the generator network. Gray states are guaranteed to be safe, while red states are inconclusive. \label{fig:safe_cells}}
% \end{figure}

The edges of the runway are at crosstrack errors of \num{-10} and \num{10} meters. Thus, any cell with a crosstrack error magnitude greater than \num{10} meters represents a failure cell and is labeled red. If the aircraft starts from any crosstrack error within the runway limits with a reasonably small heading error, the image-based controller is guaranteed to keep the aircraft on the runway. The failure states on the runway indicate states where the aircraft is near the edge of the runway and pointing away from the center line. In these states, the aircraft likely does not have enough time to change direction and therefore leaves the runway before it can change direction toward the center line.

\begin{figure*}[htb]
    \centering
    \begin{tikzpicture}[]
\begin{groupplot}[group style={horizontal sep = 1cm, vertical sep = 1.5cm, group size=5 by 1}]

\nextgroupplot [
  height = {4cm},
  ylabel = {$\theta$ (degrees)},
  title = {$t = \SI{0}{\second}$},
  xmin = {-11.0},
  xmax = {11.0},
  ymax = {30.0},
  xlabel = {$p$ (m)},
  ymin = {-30.0},
  width = {4cm},
  enlargelimits = false,
  axis on top
]

\addplot[
  ] graphics[
  xmin = -11.0,
  xmax = 11.0,
  ymin = -30.0,
  ymax = 30.0
] {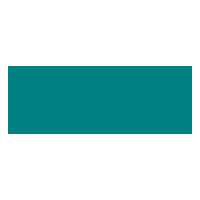};

\nextgroupplot [
  height = {4cm},
  title = {$t = \SI{5}{\second}$},
  xmin = {-11.0},
  xmax = {11.0},
  ymax = {30.0},
  xlabel = {$p$ (m)},
  ymin = {-30.0},
  width = {4cm},
  enlargelimits = false,
  axis on top
]

\addplot[
  ] graphics[
  xmin = -11.0,
  xmax = 11.0,
  ymin = -30.0,
  ymax = 30.0
] {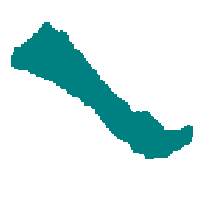};

\nextgroupplot [
  height = {4cm},
  title = {$t = \SI{10}{\second}$},
  xmin = {-11.0},
  xmax = {11.0},
  ymax = {30.0},
  xlabel = {$p$ (m)},
  ymin = {-30.0},
  width = {4cm},
  enlargelimits = false,
  axis on top
]

\addplot[
  ] graphics[
  xmin = -11.0,
  xmax = 11.0,
  ymin = -30.0,
  ymax = 30.0
] {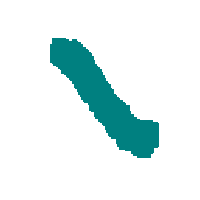};

\nextgroupplot [
  height = {4cm},
  title = {$t = \SI{15}{\second}$},
  xmin = {-11.0},
  xmax = {11.0},
  ymax = {30.0},
  xlabel = {$p$ (m)},
  ymin = {-30.0},
  width = {4cm},
  enlargelimits = false,
  axis on top
]

\addplot[
  ] graphics[
  xmin = -11.0,
  xmax = 11.0,
  ymin = -30.0,
  ymax = 30.0
] {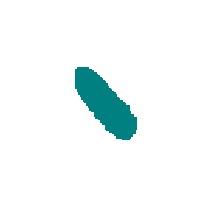};

\nextgroupplot [
  height = {4cm},
  title = {Converged},
  xmin = {-11.0},
  xmax = {11.0},
  ymax = {30.0},
  xlabel = {$p$ (m)},
  ymin = {-30.0},
  width = {4cm},
  enlargelimits = false,
  axis on top
]

\addplot[
  ] graphics[
  xmin = -11.0,
  xmax = 11.0,
  ymin = -30.0,
  ymax = 30.0
] {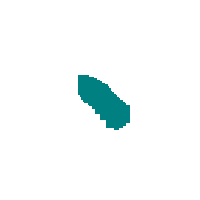};

\end{groupplot}

\end{tikzpicture}
    \caption{Overapproximated reachable set over time when starting from a state with $p \in [\SI{-10}{\meter}, \SI{10}{\meter}]$ and $\theta \in [\SI{-10}{\degree}, \SI{10}{\degree}]$. \label{fig:forward_reach}}
\end{figure*}

To test \textbf{P2}, we use the forward reachability methods from \citeauthor{Julian2019dasc}  \cite{Julian2019dasc}. The plots in \cref{fig:forward_reach} show how the overapproximated reachable set evolves over time when the aircraft starts within the limits of the runway with a heading error magnitude less than \num{10} degrees. The reachable set shrinks over time to a small region in the center of the state space. When performing forward reachability analysis, if we find that $R^\pi_{t+1}(c)= R^\pi_{t}(c)$ for all $c \in \mathcal{C}$ at given time $t$, we determine that the reachable set has converged to an invariant set. 
\begin{figure*}[htb]
    \centering
    \input{forward_reach_runway}
    \caption{Forward reachable set when starting from a state with $p \in [\SI{-8}{\meter}, \SI{8}{\meter}]$  and $\theta \in [\SI{-2}{\degree}, \SI{2}{\degree}]$ plotted on top of the simulated trajectories in \cref{fig:sims}. \label{fig:forward_reach_rw}}
\end{figure*}

In this example, the reachable set converges \num{16} seconds into the trajectory to a region of states with small heading error near the center of the runway. Thus, the neural network controller is expected to guide the aircraft to a region near the center of the runway within \num{16} seconds and keep it there for all subsequent time steps. To compare directly to the simulated trajectories, \cref{fig:forward_reach_rw} shows the overapproximated forward reachable set on top of the simulated trajectories from \cref{fig:sims} when starting from a state with $p \in [\SI{-8}{\meter}, \SI{8}{\meter}]$  and $\theta \in [\SI{-2}{\degree}, \SI{2}{\degree}]$. From this figure, we can see that the controller is guaranteed to guide the aircraft to a position near the center of the runway.

\subsection{Recall Metric}\label{sec:recall_res}
Because the guarantees we obtain through neural network verification techniques are relative to the set of images that could be generated by our generative model, it is important that our generator network adequately captures the true observation space. To test this, we solve the optimization problem in \cref{eq:generator_opt} to determine the Euclidean distance between each training image and the closest generated image. We obtained results for both the MLP simplified using supervised learning and the MLP simplified using adversarial training procedures. Because the adversarial MLP suffers from mode collapse and misses key portions of the runway (shown in \cref{fig:gen_comp}), we expect it to perform worse according to our metric.
\begin{figure}[htb]
    \centering
    \input{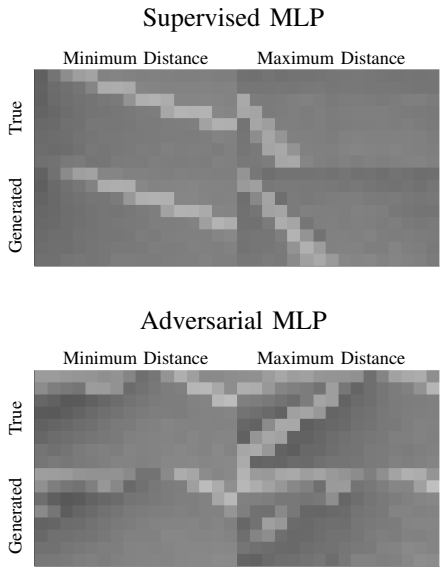}
    \caption{Closest (left) and furthest (right) generated images compared to their corresponding image in the training data for both MLPs. \label{fig:im_comp}}
\end{figure}
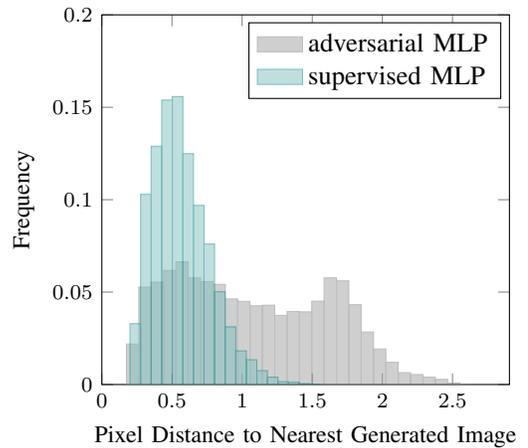
\begin{figure}[htb]
    \centering
    \begin{tikzpicture}[]
\begin{axis}[
  height = {6.5cm},
  xmin = {0.0},
  xlabel = {Pixel Distance to Nearest Generated Image},
  ylabel = {Frequency},
  yticklabel style={
        /pgf/number format/fixed,
        /pgf/number format/precision=5},
%  yticklabels = ,,,
  ymin = {0.0},
  ymax = {0.2},
  width = {7cm}
]

\addplot+[
  mark = {none},
  ybar interval,gray, fill=gray!75, opacity = 0.5, area legend
] coordinates {
  (0.1754308576832202, 219.0 / 10000.0)
  (0.2635076408512738, 527.0 / 10000.0)
  (0.35158442401932744, 554.0 / 10000.0)
  (0.43966120718738105, 617.0 / 10000.0)
  (0.5277379903554347, 664.0 / 10000.0)
  (0.6158147735234882, 577.0 / 10000.0)
  (0.7038915566915419, 557.0 / 10000.0)
  (0.7919683398595956, 542.0 / 10000.0)
  (0.8800451230276491, 463.0 / 10000.0)
  (0.9681219061957027, 450.0 / 10000.0)
  (1.0561986893637565, 426.0 / 10000.0)
  (1.14427547253181, 429.0 / 10000.0)
  (1.2323522556998636, 374.0 / 10000.0)
  (1.3204290388679172, 395.0 / 10000.0)
  (1.4085058220359707, 393.0 / 10000.0)
  (1.4965826052040245, 451.0 / 10000.0)
  (1.584659388372078, 577.0 / 10000.0)
  (1.6727361715401317, 561.0 / 10000.0)
  (1.7608129547081854, 432.0 / 10000.0)
  (1.848889737876239, 284.0 / 10000.0)
  (1.9369665210442926, 192.0 / 10000.0)
  (2.025043304212346, 120.0 / 10000.0)
  (2.1131200873803997, 64.0 / 10000.0)
  (2.2011968705484533, 54.0 / 10000.0)
  (2.2892736537165073, 40.0 / 10000.0)
  (2.377350436884561, 28.0 / 10000.0)
  (2.4654272200526144, 9.0 / 10000.0)
  (2.553504003220668, 1.0 / 10000.0)
  (2.6415807863887215, 1.0 / 10000.0)
};

\addplot+[
  mark = {none},
  ybar interval,teal, fill=teal!50, opacity = 0.5, area legend
] coordinates {
  (0.20095497075140817, 329.0 / 10000.0)
  (0.27629480492773784, 1029.0 / 10000.0)
  (0.35163463910406745, 1289.0 / 10000.0)
  (0.42697447328039706, 1539.0 / 10000.0)
  (0.5023143074567267, 1558.0 / 10000.0)
  (0.5776541416330564, 1249.0 / 10000.0)
  (0.652993975809386, 969.0 / 10000.0)
  (0.7283338099857156, 760.0 / 10000.0)
  (0.8036736441620453, 502.0 / 10000.0)
  (0.879013478338375, 312.0 / 10000.0)
  (0.9543533125147046, 182.0 / 10000.0)
  (1.0296931466910342, 134.0 / 10000.0)
  (1.1050329808673638, 75.0 / 10000.0)
  (1.1803728150436934, 40.0 / 10000.0)
  (1.255712649220023, 15.0 / 10000.0)
  (1.3310524833963528, 12.0 / 10000.0)
  (1.4063923175726825, 5.0 / 10000.0)
  (1.481732151749012, 1.0 / 10000.0)
  (1.5570719859253417, 1.0 / 10000.0)
};

\legend{adversarial MLP, supervised MLP}
\end{axis}
\end{tikzpicture}
    \caption{Distribution of distances to nearest generated image for each image in the training data for both simplified MLPs. \label{fig:radii}}
\end{figure}
\begin{figure}[htb]
    \centering
    \begin{tikzpicture}[]
\begin{axis}[
  height = {6.5cm},
  legend pos = {south east},
  ylabel = {Recall},
  xmin = {0.0},
  xmax = {2.7},
  ymax = {1.0},
  xlabel = {$r_{\max}$},
  ymin = {0.0},
  width = {7cm}
]

\addplot+[
  mark = {none},
  lightgray, very thick
] coordinates {
  (0.0, 0.0)
  (0.1, 0.0)
  (0.2, 0.0002)
  (0.3, 0.0455)
  (0.4, 0.1029)
  (0.5, 0.1743)
  (0.6, 0.2475)
  (0.7, 0.3133)
  (0.8, 0.3774)
  (0.9, 0.4363)
  (1.0, 0.4883)
  (1.1, 0.5389)
  (1.2, 0.588)
  (1.3, 0.6321)
  (1.4, 0.6754)
  (1.5, 0.7205)
  (1.6, 0.7742)
  (1.7, 0.8387)
  (1.8, 0.8972)
  (1.9, 0.939)
  (2.0, 0.9651)
  (2.1, 0.979)
  (2.2, 0.9868)
  (2.3, 0.9926)
  (2.4, 0.9969)
  (2.5, 0.9994)
  (2.6, 0.9999)
  (2.7, 1.0)
};

\addplot+[
  mark = {none},
  teal, very thick
] coordinates {
  (0.0, 0.0)
  (0.1, 0.0)
  (0.2, 0.0)
  (0.3, 0.0606)
  (0.4, 0.2151)
  (0.5, 0.4125)
  (0.6, 0.6155)
  (0.7, 0.7623)
  (0.8, 0.8694)
  (0.9, 0.931)
  (1.0, 0.9654)
  (1.1, 0.9844)
  (1.2, 0.9939)
  (1.3, 0.9979)
  (1.4, 0.9994)
  (1.5, 0.9999)
  (1.6, 1.0)
  (1.7, 1.0)
  (1.8, 1.0)
  (1.9, 1.0)
  (2.0, 1.0)
  (2.1, 1.0)
  (2.2, 1.0)
  (2.3, 1.0)
  (2.4, 1.0)
  (2.5, 1.0)
  (2.6, 1.0)
  (2.7, 1.0)
};

\legend{{}{adversarial MLP}, {}{supervised MLP}}
\end{axis}
\end{tikzpicture}
    \caption{Comparison of recall metric for the supervised and adversarial MLPs. \label{fig:recall}}
\end{figure}
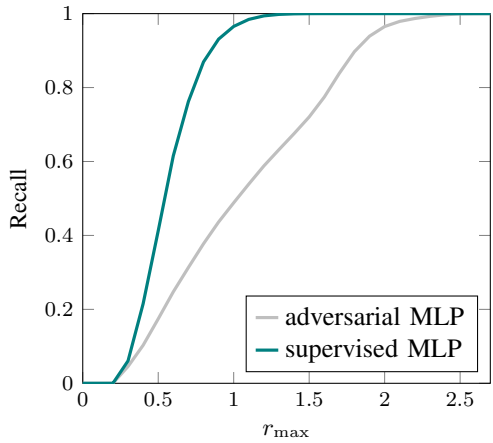

\Cref{fig:radii} shows the distribution of distances for each generator network. As expected, the supervised MLP generator has a tighter distribution centered at smaller distances indicating that it better represents the training data than the adversarial MLP. For reference, a pixel distance of one could be created by changing a single pixel in the image from white to black. In most cases, however, this error is spread out over the entire image. \Cref{fig:im_comp} compares the closest and furthest training images from the generator output space. The furthest image generated by the supervised MLP represents a slight shift in the angle at the edge line and occurs at a point near the edge of the state space. The furthest image for the adversarial MLP highlights the mode collapse by missing large sections of the centerline dash.

By selecting a maximum distance $r_{\max}$ and identifying the fraction of training images within $r_{\max}$ of a generated image, we obtain a recall metric. Varying $r_{\max}$ is equivalent to taking the cumulative distribution of the data in \cref{fig:radii}. The result of this analysis is shown in \cref{fig:recall}. This plot further highlights the improvement of the supervised MLP over the adversarial MLP. In general, this metric can be used to compare the effectiveness of generative models at capturing the training data.

\section{Conclusion}
The complicated and high-dimensional input space of image-based neural network controllers adds extra complexity to the closed-loop verification problem. In this work, we showed that we can reduce this complexity by approximating the observation space of an image-based controller using a generative model that maps a low-dimensional state and latent space to images. This insight allowed us to extend formal verification techniques previously developed for state-based controllers to image-based controllers. This method ultimately provided us with guarantees over the observation space approximated by our generative model. To quantify the validity of our approximation, we created a recall metric based on the Euclidean distance between the training images and their nearest neighbor in the generator output space.

We demonstrated our method on the autonomous aircraft taxi problem by training a GAN to produce realistic downsampled runway images. After concatenating it with the control network we wanted to verify, we showed that the controller was guaranteed to keep the aircraft on the runway over a large region of initial states. Furthermore, we showed that the controller will cause aircraft trajectories to converge to a region near the center of the runway in a finite amount of time. Finally, we calculated our recall metric for two different generator networks and found that we could use it to quantify improvements in the quality of the generator network.

This research leaves multiple avenues for future work. While we focused on semantic variations represented by the generator in this work, future work will explore combining semantic variations with noise perturbations in the verification process. Additionally, future work will apply this technique to more complicated scenarios such as other weather conditions and times of day. We could also explore the effectiveness of other deep generative models such as variational autoencoders at approximating the observation space. Although we used these methods to verify safety properties, they could also be used to find failure trajectories caused by an image-based controller. 

% Finding failures
% Other generative models
% Adaptive verification
% More complicated scenarios

\section{Acknowledgements}
The authors thank Eric Zelikman for his helpful advice throughout the progression of this work. The NASA University Leadership Initiative (grant \#80NSSC20M0163) provided funds to assist the authors with their research. This research was also supported by the National Science Foundation Graduate Research Fellowship under Grant No. DGE–1656518. Any opinion, findings, and conclusions or recommendations expressed in this material are those of the authors and do not necessarily reflect the views of any NASA entity or the National Science Foundation.

%\bibliographystyle{IEEEtran}
%\bibliography{abstract}
\renewcommand*{\bibfont}{\small}
\printbibliography

\end{document}